\documentclass[letterpaper]{article}

\usepackage{preamble/uai2020}
\usepackage[margin=1in]{geometry}

\usepackage{hyperref}

\usepackage{booktabs,multirow,multicol}       %
\usepackage{microtype}      %
\usepackage{graphicx}
\usepackage{subfigure}
\usepackage{wrapfig}
\usepackage{enumitem}

\usepackage[table]{xcolor}
\usepackage[labelfont=bf,font=footnotesize,width=.9\textwidth]{caption}
\usepackage[separate-uncertainty=true,multi-part-units=single]{siunitx}

\usepackage{adjustbox}

\usepackage{mathtools}
\usepackage{amsthm}
\usepackage{amssymb}
\usepackage{thmtools}
\usepackage{thm-restate}

\usepackage{times}

\usepackage[%
minnames=1,maxnames=99,maxcitenames=2, style=ieee, citestyle=alphabetic, doi=false, url=false,
firstinits=true,hyperref,natbib,backend=bibtex,sorting=nyt]{biblatex}%
\AtEveryBibitem{%
\ifentrytype{article}{
    \clearfield{url}%
    \clearfield{urldate}%
    \clearfield{eprint}
}{}
\ifentrytype{book}{
    \clearfield{url}%
    \clearfield{urldate}%
}{}
\ifentrytype{collection}{
    \clearfield{url}%
    \clearfield{urldate}%
}{}
\ifentrytype{incollection}{
    \clearfield{url}%
    \clearfield{urldate}%
}{}
}

\AtEveryBibitem{
    \clearfield{pages}
    \clearfield{review}%
    \clearfield{series}%
    \clearfield{volume}
    \clearfield{pages}
    \clearfield{month}
    \clearfield{eprint}
    \clearfield{isbn}
    \clearfield{issn}
    \clearlist{location}
    \clearfield{series}
    \clearlist{publisher}
    \clearname{editor}
}{}

\usepackage[capitalize]{cleveref}
\crefformat{equation}{(#2#1#3)}
\crefformat{figure}{Figure~#2#1#3}
\crefname{example}{Example}{Examples}
\crefname{lemma}{Lemma}{Lemmas}
\crefname{cor}{Corollary}{Corollaries}
\crefname{theorem}{Theorem}{Theorems}
\crefname{assumption}{Assumption}{Assumptions}

\declaretheorem[style=plain,numberwithin=section,name=Theorem]{theorem}

\declaretheorem[style=definition,sibling=theorem,name=Example]{example}
\declaretheorem[style=remark,sibling=theorem,name=Remark]{remark}

\newenvironment{example*}
 {\pushQED{\qed}\example}
 {\popQED\endexample}
\numberwithin{equation}{section}

\renewcommand{\epsilon}{\varepsilon}

\newcommand{\Reals}{\mathbf{R}}

\newcommand{\NNReals}{\Reals_{+}}

\DeclareMathOperator*{\argmin}{argmin}

\DeclareMathOperator*{\logit}{logit}

\newcommand{\EE}{\mathbb{E}}
\renewcommand{\Pr}{\mathrm{P}}
\newcommand{\convPr}{\xrightarrow{\,p\,}}

\newcommand{\given}{\mid}

\newcommand{\dist}{\ \sim\ }
\newcommand{\distiid}{\overset{\mathrm{iid}}{\dist}}

\newcommand{\cdo}{\mathrm{do}} 

\newcommand{\qhat}{$\hat{\psi}^{Q}$}
\newcommand{\plugin}{$\hat{\psi}^{\mathrm{plugin}}$}

\providecommand\given{} 
\newcommand\SetSymbol[1][]{
  \nonscript\,#1:\nonscript\,\mathopen{}\allowbreak}
\DeclarePairedDelimiterX\Set[1]{\lbrace}{\rbrace}%
{ \renewcommand\given{\SetSymbol[]} #1 }

\newcommand{\grad}{\nabla}
\DeclareRobustCommand{\parhead}[1]{\textbf{#1}~}
\newcommand{\maxf}[1]{{\cellcolor[gray]{0.8}} #1}
\global\long\def\embedding{\lambda}
\global\long\def\globparam{\gamma}

\usepackage{xcolor}

\definecolor{WowColor}{rgb}{.75,0,.75}
\definecolor{SubtleColor}{rgb}{0,0,.50}

\newcounter{margincounter}

\newcommand*\samethanks[1][\value{footnote}]{\footnotemark[#1]}
\title{Adapting Text Embeddings for Causal Inference}
\author{ {\bf Victor Veitch\thanks{~~Equal contribution.}} \\
\And
{\bf Dhanya Sridhar\samethanks}  \\
\\
Department of Statistics and Department of Computer Science\\
Columbia University
\And
{\bf David M. Blei}   \\
}

\begin{document}

\maketitle

\begin{abstract}
Does adding a theorem to a paper affect its chance of
acceptance? Does labeling a post with the author's gender
affect the post popularity?  This paper develops a method to
estimate such causal effects from observational text data,
adjusting for confounding features of the text such as the
subject or writing quality. We assume that the text suffices
for causal adjustment but that, in practice, it is
prohibitively high-dimensional.  To address this challenge, we
develop \textit{causally sufficient embeddings},
low-dimensional document representations that preserve
sufficient information for causal identification and allow for
efficient estimation of causal effects. Causally sufficient
embeddings combine two ideas.  The first is supervised
dimensionality reduction: causal adjustment requires only the
aspects of text that are predictive of both the treatment and
outcome.  The second is efficient language modeling:
representations of text are designed to dispose of
linguistically irrelevant information, and this information is
also causally irrelevant.  Our method adapts language models
(specifically, word embeddings and topic models) to learn
document embeddings that are able to predict both treatment
and outcome.  We study causally sufficient embeddings with
semi-synthetic datasets and find that they improve causal
estimation over related embedding methods.  We illustrate
the methods by answering the two motivating questions---the
effect of a theorem on paper acceptance and the effect of a
gender label on post popularity.
Code and data available at \href{https://github.com/vveitch/causal-text-embeddings-tf2}{github.com/vveitch/causal-text-embeddings-tf2}.
\end{abstract}

\section{INTRODUCTION}

This paper is about causal inference on text.
\begin{example}
  Consider a corpus of scientific papers submitted to a conference.
  Some have theorems; others do not.  We want to infer the causal
  effect of including a theorem on paper acceptance.  The effect is
  confounded by the subject of the paper---more technical topics
  demand theorems, but may have different rates of acceptance.  The
  data does not explicitly list the subject, but it does include each
  paper's abstract.  We want to use the text to adjust for the subject
  and estimate the causal effect.
\end{example}

\begin{example}
  Consider comments from Reddit.com, an online forum. Each post has a
  popularity score and the author of the post may (optionally) report
  their gender.  We want to know the direct effect of a `male' label
  on the score of the post.  However, the author's gender may
  affect the text of the post, e.g., through tone, style, or topic
  choices, which also affects its score.  Again, we want to use the
  text to accurately estimate the causal effect.
\end{example}

In these two examples, the text encodes features such as the subject
of a scientific paper or the writing quality of a Reddit comment. 
These features bias the estimation of causal effects from observed
text documents.
By assumption, the text carries sufficient information to identify the causal effect;
we can use adjustment methods from causal inference to estimate
the effects.
But in practice we have finite data and the text is high
dimensional, prohibiting efficient causal inference.  The
challenge is to reduce the text to a low-dimensional representation
that suffices for causal identification and enables efficient
estimation from finite data.
We refer to these text representations as \textit{causally sufficient embeddings}.

The method for learning causally sufficient embeddings 
is based on two ideas.
The first comes from examining causal adjustment.
Causal adjustment only requires the parts of text
that are predictive of the treatment and outcome.
Causally sufficient embeddings preserve such predictive
information while discarding the parts of text that are
irrelevant for causal adjustment.
We use supervised dimensionality
reduction to learn embeddings that 
predict the treatment and outcome.

The second idea for learning the embeddings
comes from research on language modeling
\citep[e.g.,][]{Mikolov:Chen:Corrado:Dean:2013,Mikolov:Sutskever:Chen:Corrado:2013,Devlin:Chang:Lee:Toutanova:2018,Peters:Neumann:Iyyer:Gardner:Clark:Lee:Zettlemoyer:2018}.
Outcomes such as paper acceptance or comment popularity
are judgments made by humans, and human judgments stem from
processing natural language. 
Thus, when performing causal adjustment for text, the confounding
aspects must be linguistically meaningful.
Consequently, causally sufficient embeddings model the language
structure in text.

We combine these two ideas to adapt
modern methods for language modeling --- BERT
\cite{Devlin:Chang:Lee:Toutanova:2018} and topic models \cite{blei2003latent}
--- in service
of causal inference.
Informally, we learn embeddings of text documents
that retain only information that is predictive of the treatment and outcome,
and relevant for language understanding.

We empirically study these methods.
Any empirical evaluation must use semi-synthetic data since ground truth causal effects are not available in real data. 
A key challenge is that text is hard to simulate, 
and realistic models that explicitly relate text to confounding aspects are not available.
We show how to circumvent these issues by using real text documents and extra observed features of the documents as confounders.
The empirical evaluation demonstrates the advantages of supervised dimensionality
reduction and language modeling for producing
causally sufficient embeddings.
Code and data to reproduce the studies will be publicly available.

\parhead{Contributions.} The contributions of this paper are the following: 
1) Adapting modern text representation learning methods to estimate causal effects
from text documents; 
2) Establishing the validity of this estimation procedure;
3) A new approach for empirically validating causal estimation problems with text, based on semi-synthetic data.

\parhead{Related work.}  This paper connects to several areas of
related work.

The first area is causal inference for text.
\citet{roberts2018adjusting} also discuss how to estimate effects of
treatments applied to text documents.  They rely (in part) on topic
modeling to reduce the dimension of the text.  
In this paper, we adapt topic modeling
to produce representations that predict the treatment and outcome
well, and demonstrate empirically that this strategy
improves upon unsupervised topic modeling for causal effect estimation.

In other work, \citet{egami2018make} reduce raw text
to interpretable outcomes; \citet{wood2018challenges} estimate
treatment effects when confounders are observed, but missing or noisy
treatments are inferred from text.  
In contrast, we are concerned with text as the confounder.

A second area of related work addresses causal inference with
unobserved confounding when there is an observed proxy for the
confounder \cite{Kuroki:Miyakawa:1999, Pearl:2012, Kuroki:Pearl:2014,
  Miao:Geng:TchetgenTchetgen:2018,kallus2018causal}.  They
usually assume that the observed proxy variables are noisy
realizations of the unobserved confounder, and then derive conditions
under which causal identification is possible.  One view of our
problem is that each unit has a latent attribute (e.g., topic) such
that observing it would suffice for causal identification, and the
text is a proxy for this attribute.  Unlike the proxy variable
approach, however, we assume the text fully captures confounding. Our
interest is in methods for finite-sample estimation rather than
infinite-data identification.

Finally, there is work on
adapting representation learning for effect estimation
by directly optimizing causal criteria such as balance and overlap 
\cite{johansson2016learning,johansson2018learning,johansson2019support,johansson2020generalization,dcovariate}.
As in this paper, \citet{shi2019adapting,veitch2019using} learn representations
by predicting the treatment and outcome. 
We extend these ideas to
learn text representations.

\section{BACKGROUND}
\label{sec:setup}
We begin by fixing notation and recalling some ideas from the
estimation of causal effects.  Each statistical unit is a document
represented as a tuple $O_i=(Y_{i},T_{i}, \mathbf{W}_{i})$, where
$Y_{i}$ is the outcome, $T_{i}$ is the treatment, and $\mathbf{W}_i$
is the sequence of words.  The observed dataset contains $n$
observations drawn independently and identically at random from a
distribution, $O_{i} \sim P$.

\begin{figure}
	\centering
	\includegraphics[width=6cm]{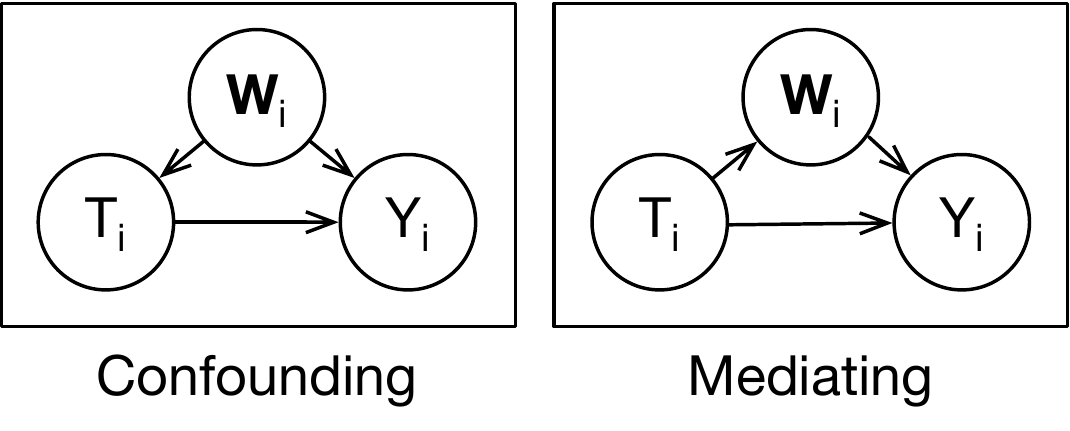}
	\caption{Models for the ATT (left) and NDE (right). \label{fig:causalgraph}}
\end{figure}

We review estimation of the average treatment effect on the treated (ATT) and the natural
direct effect (NDE).  For both, we assume that the words are sufficient for
adjustment.

\parhead{ATT.}  The ATT is
\[
\psi=\EE[Y \given \cdo(T=1), T=1]-\EE[Y\given \cdo(T=0), T=1].
\]
Pearl's $\cdo$ notation indicates that the effect of
interest is causal: what happens if we \emph{intervene} by adding a
theorem to a paper, given that we observe that it has a theorem?
We assume that the words $\mathbf{W}_i$ carry
sufficient information to adjust for confounding (common causes)
between $T_i$ and $Y_i$. \Cref{fig:causalgraph} on the left depicts this assumption.  
 We define $Z_{i} = f(\mathbf{W}_i)$ to be
the part of $\mathbf{W}_i$ which blocks all
`backdoor paths' between $Y_{i}$ and $T_{i}$.
The causal effect is then identifiable from observational data as:
\begin{equation}
\psi=\EE[\EE[Y\given Z,T=1]-\EE[Y\given Z,T=0]\given T=1].\label{eq:psi_def}
\end{equation}

Our task is to estimate the ATT $\psi$ from a finite data sample.
Define $Q(t,z)=\EE[Y\given t,z]$ to be the conditional expected
outcome and $\hat{Q}$ to be an estimate for $Q$. 
Define $g(z) = \Pr(T=1\given z)$ to be the propensity score and
$\hat{g}$ to be an estimate for $g$.  The ``plugin'' estimator of
\ref{eq:psi_def} is
\begin{equation}\label{eq:psiq}
\hat{\psi}^{\textrm{plugin}} = \frac{1}{n} \sum_i
\left[\hat{Q}(1,z_{i})-\hat{Q}(0,z_{i})\right] \hat{g}(z_i) /
\left(\frac{1}{n} \textstyle \sum_i t_i\right).
\end{equation}
Here $\psi$ is estimated by a two-stage procedure: First estimate
$\hat{Q}$ and $\hat{g}$ with predictive models; then plug $\hat{Q}$
and $\hat{g}$ into a pre-determined statistic to compute the estimate
of the ATT.  What is important is that the estimator depends on $z_i$
only through $\hat{Q}(t,z_i)$ and $\hat{g}(z_i)$.

The $Q$-only estimator, which only uses the conditional expected outcomes, is
\begin{equation}\label{eq:psiqonly}
\hat{\psi}^{Q} =\frac{1}{\textstyle \sum_i t_i} \sum_i
t_i\left[\hat{Q}(1,z_{i})-\hat{Q}(0,z_{i})\right].
\end{equation}

\parhead{NDE.}  
The direct effect is the expected
change in outcome if we apply the treatment while holding fixed any
mediating variables that are affected by the treatment and that affect
the outcome. 
\Cref{fig:causalgraph} on the right depicts the text as mediator
of the treatment and outcome.
For the estimation of the direct effect, we take
$Z = f(\mathbf{W})$ to be the parts of $\mathbf{W}_i$ that mediate $T$ and $Y$. 
 The natural direct effect of
treatment $\beta$ is average difference in outcome induced by giving
each unit the treatment, if the distribution of $Z$ had been as though
each unit received treatment.  That is,
\[
\beta =\EE_{Z|T=1}[\EE[Y \given Z, \cdo(T=1)] - \EE[Y \given Z, \cdo(T=0)]].
\]
In the Reddit example, this is the expected difference in score between a
post labeled as written by a man versus labeled as written by a woman,
where the expectation is taken over the distribution of posts written by men.

Mathematically, $\beta$ is equivalent to $\psi$ in \eqref{eq:psi_def}.
The causal parameters have different interpretations depending on the graph.
Under minimal conditions, the NDE may be estimated from observational data \cite{Pearl:2014}.
The estimators for $\beta$ are the same 
as those for the ATT, given in \eqref{eq:psiq} and \eqref{eq:psiqonly}
\citep[][Ch.~8]{vanderLaan:Rose:2011}.

\section{CAUSAL TEXT EMBEDDINGS}
\label{sec:method}

Following the previous section, we want to produce estimates of the
propensity score $g(z_{i})$ and the conditional expected outcome
$Q(t_{i},z_{i})$.
We assume that some properties of the text $z_i = f(\mathbf{w}_i)$
suffice for identification.
These properties are generally lower-dimensional than the text
$\mathbf{w}_i$ itself and are linguistically meaningful (i.e., they have to do with what the language means)
But, we do not directly observe
the confounding features $z_i$.  Instead, we observe the raw text.

A simple approach is to abandon $z_i$ and learn models for the
propensities and conditional outcomes directly from the words
$\mathbf{w}_i$.  Since $\mathbf{w}_i$ contains all information about
$z_i$, the direct adjustment will also render the causal effect
identifiable.  Indeed, in an infinite-data setting this approach would
be sound.  However, the text is high-dimensional and with finite data,
this approach produces a high-variance estimator.

To this end, our goal is to reduce the words $\mathbf{w}_i$ to a
feature $z_i$ that contains sufficient information to render the
causal effect identifiable and that allows us to efficiently learn the
propensity scores and conditional outcomes with a finite sample of
data.

Our strategy is to use the words of each document to produce an
embedding vector $\lambda(\mathbf{w})$ that captures the confounding
aspects of the text.  
Using the embedding, the propensity score is
 is $g(\lambda(\mathbf{w}))=\Pr(T=1 \given \lambda(\mathbf{w}))$.
The conditional outcomes are
$Q(t,\lambda(\mathbf{w}))=\EE[Y\given t,\lambda(\mathbf{w})]$.
The embeddings $\lambda(\mathbf{w})$ are causally sufficient if we
can use them to estimate the propensities and conditional outcomes
required by the downstream effect estimator.
This result builds on \cite{ROSENBAUM:RUBIN:1983}; \Cref{thm:prediction-suffices} makes this formal.

In finite sample, additional domain knowledge is
useful for finding these embeddings more efficiently.  With text in
particular, we assume that features that are useful for language understanding
are also useful for eliminating confounding.  
The reason is that
humans interpret the language and then produce the outcome based
on aspects such as topic, writing quality or sentiment.

To produce causally sufficient embeddings, we will adapt models for language understanding
and refine the representations they produce to predict the treatment and outcome.
Informally, these models 
take in words $\mathbf{w}_i$ and produce a
tuple
$(\lambda_i,\tilde{Q}(t_{i},\lambda_{i}), \tilde{g}(\lambda_{i}))$,
which contains an embedding $\lambda_i$ and estimates of $g$ and $Q$
that use that embedding.  
Such models provide an
effective black-box tool for both distilling the words into the
information relevant to prediction problems, and for solving those
prediction problems.

Finally, to estimate the average treatment effect, we follow the strategy of \cref{sec:setup}.  
Fit the
embedding-based prediction model to produce estimated embeddings
$\hat{\lambda}_i$, propensity scores $\tilde{g}(\hat{\lambda}_{i})$
and conditional outcomes
$\tilde{Q}(t_{i},\hat{\lambda}_{i})$.  Then, plug these values into
a downstream estimator.  

We develop two methods that produce causally sufficient document
embeddings: causal BERT and the causal topic model.

\parhead{Causal BERT.}
\global\long\def\samp{\mathsf{Sample}}
\global\long\def\crossent{\mathsf{CrossEnt}}
We modify BERT, a state-of-the-art language model \citep{Devlin:Chang:Lee:Toutanova:2018}.
Each input to BERT is the document text, a sequence of word-piece tokens
$\mathbf{w}_i=(w_{i1},\dots,w_{il})$. 
The model is tasked with producing three kinds of outputs: 1) document-level embeddings, 2) a map from the embeddings to treatment probability, 3)
a map from the embeddings to expected outcomes for the treated and untreated.

The model assigns an embedding $\xi_w$ to each word-piece $w$.
It then produces a document-level embedding for document text $\mathbf{w}_i$ as 
$\embedding_i = f((\xi_{w_{i1}},\dots,\xi_{w_{il}}), \globparam^{{\text{U}}})$ for a particular function $f$.
The embeddings and global parameter $\globparam^{\text{U}}$ are trained by minimizing an unsupervised objective, denoted as $L_{\text{U}}(\mathbf{w}_i; \xi, \globparam^{\text{U}})$.
Informally, random word-piece tokens are censored from each document and the model is tasked with predicting their identities.\footnote{BERT also considers a `next sentence' prediction task, which we do not use.}

Following \citet{Devlin:Chang:Lee:Toutanova:2018}, we use a
fine-tuning approach to solve the prediction problem.  
We add a logit-linear layer mapping
$\lambda_i \to \tilde{g}(\lambda_{i}; \globparam^g)$ and a 2-hidden
layer neural net for each of
$\lambda_i \to \tilde{Q}(0,\lambda_{i}; \globparam^{Q_0})$ and
$\lambda_i \to \tilde{Q}(1,\lambda_{i}; \globparam^{Q_1})$.  We learn
the parameters for the embedding model and the prediction model
jointly.  
This supervised dimensionality reduction adapts the 
embeddings to be useful for the
downstream prediction task, i.e., for causal inference.

We write $\gamma$ for the full collection of global parameters. The final model is trained as:
\begin{align*}
\hat{\embedding}_{i} &= f((\hat{\xi}_{n,w_{i1}},\dots,\hat{\xi}_{n,w_{il}}), \hat{\globparam}^{\text{U}}) \\
\hat{\xi},\hat{\globparam} &=\argmin_{\xi,\globparam}\frac{1}{n}\sum_i L(\mathbf{w}_i; \xi, \globparam),
\end{align*}
where the objective is designed to predict both the treatment and
outcome.  It is
\begin{align*}
\begin{split}
L(\mathbf{w}_i;\xi,\globparam) = &\big(y_{i} - \tilde{Q}(t_i, \embedding_i; \globparam) \big)^2\\ 
&+ \crossent \big(t_{i},\tilde{g}(\embedding_i; \globparam) \big)\\
&+ L_{\text{U}}(\mathbf{w}_i; \xi, \globparam).
\end{split}
\end{align*}

\parhead{Effect estimation.} Computing causal effect estimates simply
requires plugging in the propensity scores and expected outcomes that
the trained model predicts on the held-out units.  For example, using
the plug-in estimator \cref{eq:psiq},
\begin{equation}
\label{eq:psiq2}
\hat{\psi}^Q := \frac{1}{n} \sum_{i}\tilde{Q}(1,\hat{\embedding}_{n,i};\hat{\globparam}_{n}^{Q})-\tilde{Q}(0,\hat{\embedding}_{n,i};\hat{\globparam}_{n}^{Q}).
\end{equation}
The estimation procedure is the same for the NDE.

\parhead{Causal Amortized Topic Model.}
We adapt the standard topic model \cite{blei2003latent}, a generative model of 
text documents.
Formally, a document $\mathbf{w}_i$ (a sequence of word tokens) is generated
from $k$ latent topics by:
\begin{align*}
r_i &\sim \mathcal{N}(0, I)\\
\theta_i &= \textrm{softmax}(r_i) \\
w_{ij} &\sim \textrm{Cat}(\theta_i^\top \beta)
\end{align*}
The vector $\theta_i$ represents the document's topic proportions, drawn
from log normal prior distribution. 
It is the embedding $\lambda(\mathbf{w}_i)$ of the document.
The parameters $\beta$ represent topics.

Typically, this model is fit with variational inference, which seeks to maximize
a lower bound of the marginal log likelihood of the documents. It is a sum
of per-document bounds. Each is
\begin{equation}
\mathcal{L}_i(\beta, \eta) = \mathbb{E}_{q(\theta; \eta)}[p(\mathbf{w}_i \given \theta_i;\beta)]
- \mathcal{D}_{\textrm{KL}}(q(\theta_i;\eta), p(\theta_i)). \nonumber
\end{equation}
The first term pushes the variational distribution over document representations $q(\theta;\eta)$ to reconstruct the observed documents well. This term encourages
language modeling.

We use amortized inference to define a variational family $q(\theta \given \mathbf{w}; \eta)$ that depends on the observed documents. It uses a feedforward neural network called an encoder with parameters $\eta$ to produce a representation $\theta$. For each document, the encoder produces two vectors $(\mu_i, \Sigma_i)$ such that 
$q(\theta_i \given \mathbf{w}; \eta) = \mathcal{N}(\mu_i, \Sigma_i)$. 
This is the amortized topic model (ATM).

We adapt the training objective of this model
to produce representations that predict the treatment and outcome well.
A logit linear mapping $\theta_i \rightarrow \tilde{g}(\theta_i;\gamma^g)$ produces propensity scores from the document representation.
A linear mapping $\theta_i \rightarrow \tilde{Q}(t_i, \theta_i;\gamma^Q)$ produces 
expected outcomes from the document representations.
The final loss is
\begin{align*}
\begin{split}
L(\mathbf{w}_i; \eta, \beta, \gamma) &= -\mathcal{L}_i(\beta, \eta) \\
&+ \mathbb{E}_{q(\theta | \mathbf{w}; \eta)}[\crossent \big(t_{i},\tilde{g}(\theta_i;\gamma)]\\
&+ \mathbb{E}_{q(\theta | \mathbf{w}; \eta)}[y_i - \tilde{Q}(t_i, \theta_i; \gamma)^2].
\end{split}
\end{align*} 
It encourages good reconstruction of both the observed documents,
and the treatment and outcome under the learned representation.
The objective is minimized with stochastic optimization, forming noisy gradients
using the reparameterization trick. 
We refer to this model as the causal amortized topic model (Causal ATM).

\parhead{Validity.} 
The central idea of the method is that instead of adjusting for
all of the information in $\mathbf{w}$ it suffices to adjust for
the limited information in $\lambda(\mathbf{w})$.
We now formalize this observation and establish the validity
of our estimation procedure.
To avoid notational overload, we state the result for only the ATT.
The same arguments carry through for the NDE as well.

A key observation is that for $z$ to be confounding,
it must causally influence both treatment assignment and the outcome.
Accordingly, $z$ must be predictive of both the treatment and outcome.
Put differently: any information in $\mathbf{w}$ that is not predictive
for both $T$ and $Y$ is not relevant for $z$, is not confounding, and may be safely excluded from
$\lambda(\mathbf{w})$.
Or, if $\lambda(\mathbf{w})$ carries the relevant information for the prediction
task then it is also causally sufficient.
The next result formalizes the observation that predictive sufficiency is also causal sufficiency.
\begin{theorem}\label{thm:prediction-suffices}
  Suppose $\lambda(\mathbf{w})$ is some function of the words such that
  at least one of the following is $\lambda(\mathbf{W})$-measurable:
  \begin{enumerate}[nosep]
  \item $(Q(1,\mathbf{W}), Q(0,\mathbf{W}))$,
  \item $g(\mathbf{W})$,
  \item $g((Q(1,\mathbf{W}), Q(0,\mathbf{W})))$ or $(Q(1,g(\mathbf{W})), Q(0,g(\mathbf{W})))$.
  \end{enumerate}
  If adjusting for $\mathbf{W}$ suffices to render the ATT identifiable then
  adjusting for only $\lambda(\mathbf{W})$ also suffices.
  That is, $\psi=\EE[\EE[Y\given \lambda(\mathbf{W}), T=1]-\EE[Y\given \lambda(\mathbf{W}), T=0]]$.
\end{theorem}
In words: the random variable $\lambda(\mathbf{W})$ carries the
information about $\mathbf{W}$ relevant to the prediction of both the
propensity score and the conditional expected outcome.  While
$\lambda(\mathbf{W})$ will typically throw away much of the
information in the words, \cref{thm:prediction-suffices} says that
adjusting for it suffices to estimate causal effects.
Item 1 is immediate from \cref{eq:psi_def}, and says it suffices to preserve information about $Y$.   
Item 2 is a classic result of \citep[][Thm.~3]{ROSENBAUM:RUBIN:1983}, and says it suffices to preserve information about $T$.
Item 3 weakens our requirements further: we can even throw away information relevant to $Y$, so
long as this information is not also relevant to $T$ (and vice versa).

The next result extends this identification result to estimation,
establishing conditions for the validity of our estimation procedure.
\begin{theorem}\label{thm:validity}
	Let $\eta(z) = (\EE[Y \! \given\! T\!=\!0, z], \EE[Y \! \given\! T\!=\!1, z], \Pr[T \!= \! 1 \! \given\! z))$ be the
	conditional outcomes and propensities given $z$.
	Suppose that $\hat{\psi}(\{(t_i, y_i, z_i)\}; \eta) = \frac{1}{n} \sum_i \phi(t_i, y_i, \eta(z_i)) + o_p(1)$ is some consistent estimator for the average treatment effect $\psi$.
	Further suppose that there is some function $\lambda$ of the words such that 
	\begin{enumerate}[nosep]
		\item (identification) $\lambda$ satisfies the condition of \cref{thm:prediction-suffices}.
		\item (consistency) $\|\eta(\lambda(\mathbf{W}_i)) - \tilde{\eta}(\hat{\lambda}_{i})\|_{P,2} \to 0$ as $n \to \infty$,
		where $\tilde{\eta}$ is the estimated conditional outcome and propensity model.
		\item (well-behaved estimator) $\|\grad_\eta \phi(t,y,\eta)\|_2 \le C$ for some constant $C \in \NNReals$,
	\end{enumerate}
	then, $\tilde{\psi}(\{(t_i, y_i, \hat{\lambda}_i)\}; \tilde{\eta}) \convPr \psi$.
\end{theorem}
\begin{remark}
	The requirement that the estimator $\hat{\psi}$ behaves asymptotically as a sample mean is not an important restriction;
	most commonly used estimators have this property \cite{Kennedy:2016}.
	The third condition is a technical requirement on the estimator. In the cases we consider, it suffices that the range of $Y$ and $Q$ are bounded and that $g$ is bounded away from 0 and 1. This later requirement is the common `overlap' condition, and is anyway required for the estimation of the causal effects.
\end{remark}
The proof is given in the appendix.

As with all causal inference, the validity of the procedure relies on uncheckable assumptions that the practitioner must
assess on a case-by-case basis. Particularly, we require that:
\begin{enumerate}
	\item (properties $z$ of) the document text renders the effect identifiable,
	\item the embedding method extracts semantically meaningful text information relevant to the prediction of both $t$ and $y$,
	\item the conditional outcome and propensity score models are consistent.
\end{enumerate}
Only the second assumption is non-standard.  In practice, we use the
best possible embedding method and take the strong performance on
(predictive) natural language tasks in many contexts as evidence that
the method effectively extracts information relevant to prediction
tasks.  
Additionally, we use the domain knowledge
that features which are useful for language understanding
are also relevant for adjusting for confounding.
Informally, assumption 2 is satisfied if we use a good
language model, so we satisfy it by using the best available
model.

\section{EXPERIMENTS}
\label{sec:experiments}

We now empirically study the causally sufficient embeddings
produced by causal BERT and causal ATM.\footnote{Code and data available at \href{https://github.com/vveitch/causal-text-embeddings-tf2}{github.com/vveitch/causal-text-embeddings-tf2}}
The question of interest is whether supervised
dimensionality reduction and language modeling
produce embeddings that admit efficient causal estimation.

Empirically validating effect estimation is difficult since known causal
effects in text are unavailable. We address this gap with
 semi-synthetic data
 We use real text documents and simulate an outcome that depends on both the treatment
of interest and a confounder. 

We find: 1) modeling language improves effect estimation;
2) supervised representations perform better than their unsupervised counterparts;
3) the causally sufficient representations proposed in this paper effectively adjust for confounding.

Finally, we apply causal BERT to the two
motivating examples in the introduction. We estimate causal effects on
paper acceptance and post popularity on Reddit.com.
Our application suggest that much of the apparent
treatment effects is attributable to confounding in the text.
Code and data to reproduce all studies will be publicly available.

\parhead{PeerRead.} PeerRead is a corpus of computer-science papers
\cite{Kang:Ammar:Dalvi:vanZuylen:Kohlmeier:Hovy:Schwartz:2018}.  We
consider a subset of the corpus consisting of papers posted to the
arXiv under \texttt{cs.cl}, \texttt{cs.lg}, or \texttt{cs.ai} between
2007 and 2017 inclusive.  The data only includes papers which are not
cross listed with any non-\texttt{cs} categories and are within a
month of the submission deadline for a target conferences.  The
conferences are: ACL, EMNLP, NAACL, EACL, TACL, NeurIPS, ICML, ICLR
and AAAI.  A paper is marked as accepted if it appeared in one of the
target venues.  Otherwise, the paper is marked as rejected.  The
dataset includes 11,778 papers, of which 2,891 are accepted.

For each paper, we consider the text of abstract,
the accept/reject decision, and two attributes that can be predicted from the abstract text:
\begin{enumerate}
	\item \texttt{buzzy}: the title contains any of `deep', `neural', `embed', or `adversarial net'.
	\item \texttt{theorem}: the word `Theorem' appears in the paper.
\end{enumerate}

\parhead{Reddit.} Reddit is an online forum divided into
topic-specific subforums called `subreddits'.  We consider three
subreddits: \texttt{keto}, \texttt{okcupid}, and \texttt{childfree}.
In these subreddits, we identify users whose username flair includes a
gender label (usually `M' or `F').  We collect all top-level
comments %
from these users in 2018.
We use each comment's text and score, the number of likes minus dislikes
from other users.
The dataset includes 90k comments in the selected subreddits. We
consider the direct effect of the labeled gender on posts' scores.

\subsection{Empirical Setup}
Empirical evaluation of causal estimation procedures requires semi-synthetic data
because ground truth causal effects are usually not available.
We want semi-synthetic data to be reflective of the real world.
This is challenging in the text setting: it is difficult to generate
text on the basis of confounding aspects such as topic or writing quality.
We use real text and metadata ---subreddit and title buzziness---as the confounders $\tilde{z}$ for the simulation. 
We checked that these confounders are related to the text.
We simulate only the outcomes, using the treatment and the confounder.
We compute the true propensity score $\pi(\tilde{z})$
as the proportion of units with $t_i=1$ in each strata of $\tilde{z}$.
Then, $Y_i$ is simulated from the model:
\[
Y_i = t_i + b_1(\pi(\tilde{z_i}) - 0.5) + \epsilon_i \qquad	\epsilon_i \sim N(0, \gamma).
\]
Or, for binary outcomes, 
\[ Y_i \sim \textrm{Bernoulli}(\sigma(0.25t_i + b_1(\pi(\tilde{z_i}) - 0.2)))\]
The parameter $b_1$ controls the level of confounding; e.g.,
the bias of the unadjusted difference $\EE[Y|T=1]  - \EE[Y|T=0]$ increases with $b_1$.
For PeerRead, we report estimates of the ATE for binary simulated outcomes. 
For Reddit, we compute the NDE for simulated real-valued outcomes.

\parhead{Methods.}
The goal is to study the utility of language modeling and supervision in representation
learning. We use Causal BERT (C-BERT) and Causal ATM (C-ATM), explained in \cref{sec:method}, to test 
these ideas. 

First, we compare the methods to two that omit language modeling. 
We fit C-BERT and C-ATM without the loss terms that encourage
language modeling: in BERT, it is the censored token prediction and in ATM,
it is the expected reconstruction loss. 
They amount to fitting attention-based or multi-layer feedforward supervised
models. We refer to these baselines as BERT (sup. only) and NN.

Second, we consider methods that omit supervised representation learning.
These include fitting regression models for the expected outcomes and propensity scores with fixed representations of documents. We compare four: 1) bag-of-words (BOW), 2) out-of-the-box BERT
embeddings, 3) the document-topic proportions produced by latent Dirichlet allocation (LDA), 4) and the document representation produced by ATM. 
To produce document-level embeddings from BERT, the embeddings
for each token in the document are summed together.

\parhead{Estimator.} 
For each experiment, we consider two downstream estimators:
the plugin estimator \cref{eq:psiq} and the $Q$-only estimator \cref{eq:psiqonly}.
For all estimators, we exclude units that have a
predicted propensity score greater than 0.97 or less than 0.03.

\parhead{BERT pre-processing.}
For BERT models, we truncate PeerRead abstracts to 250 word-piece
tokens, and Reddit posts to 128 word-piece tokens.  We begin with a
BERT model pre-trained on a general English language corpus.  We
further pre-train a BERT model on each dataset, running training on
the unsupervised objective until convergence.  

\begin{table*}[h!]
  \caption{
  	Comparisons on semi-simulated data show that effect estimation is improved by: (a) language modeling (left); (b)supervised dimensionality reduction (right).
  	(a) C-BERT and C-ATM improve effect estimation over NN and BERT (sup. only), which do not model language structure. 
  	(b) C-BERT and C-ATM improve effect estimation over BOW, BERT, LDA and ATM, which produce representations
  	that are not supervised to predict the treatment and outcome.
  	The tables report estimated NDE for Reddit and estimated ATT for PeerRead.
  	Shaded numbers are closest to the ground truth.
    The simulation setting used is  $\beta_1=10.0, \gamma=1.0$ for Reddit
    and  $\beta_1=5.0$ for PeerRead. }\label{tb:q1andq2}
  \begin{center}
    \begin{tabular}{lcc}
      \multicolumn{3}{c}{\textbf{(a) Language Modeling Helps}}\\
     \toprule
      \multicolumn{1}{r}{Dataset:\ \ }  & Reddit & PeerRead \\
      & (NDE) & (ATT) \\
      \midrule
	Ground truth & 1.00 & 0.06 \\
	Unadjusted & 1.24 & 0.14 \\
	\cmidrule{1-3}
 	NN \qhat & 1.17  & 0.10  \\
 	NN \plugin &  1.17 & 0.10 \\ 
	BERT (sup. only)  \qhat &0.93 & 0.19 \\
	BERT (sup. only)  \plugin & 1.17 & 0.18 \\
	\textbf{C-ATM} \qhat  & 1.16 & 0.10 \\
	\textbf{C-ATM} \plugin & 1.13 & 0.10 \\
	\textbf{C-BERT }\qhat & \maxf{1.07} & \maxf{0.07}\\
	\textbf{C-BERT} \plugin & 1.15 & 0.09 \\
      \bottomrule
    \end{tabular}
		\begin{tabular}{lcc}
	\multicolumn{3}{c}{\textbf{(b) Supervision Helps}}\\
	\toprule
	\multicolumn{1}{r}{Dataset:\ \ }  & Reddit & PeerRead\\
	      & (NDE) & (ATT) \\
	\midrule
	Ground truth & 1.00 & 0.06 \\
	Unadjusted & 1.24 & 0.14 \\
	\cmidrule{1-3}
	BOW \qhat & 1.17 & 0.13\\
	BOW \plugin & 1.18 & 0.14\\ 
	BERT  \qhat & -15.0 & -0.25 \\
	BERT  \plugin & -14.1 & -0.28 \\
	LDA \qhat & 1.20 & 0.07 \\
	LDA \plugin  & 1.20 & 0.09 \\
	ATM \qhat & 1.17 & 0.08 \\
	ATM \plugin &  1.17 & 0.08\\ 
	\bottomrule
\end{tabular}
  \end{center}
\end{table*}

\begin{table*}[h!t]
	\caption{Embedding adjustment recovers the NDE on Reddit.
		This persists even with high confounding and high noise.
		Table entries are estimated NDE.
		Columns are labeled by confounding level.
		Low, Med., and High correspond to $\beta_1=1.0,10.0$ and $100.0$.  }\label{tb:simReddit}
	\begin{center}
		\begin{tabular}{l @{\hskip 3em} *{3}{S} @{\hskip 3em} *{3}{S}}
			\toprule
			\multicolumn{1}{r}{Noise:\ \ } & \multicolumn{3}{c}{$\gamma=1.0$ } & \multicolumn{3}{c}{$\gamma=4.0$}\\
			
			\multicolumn{1}{r}{Confounding:\ \ }  & {Low} & {Med.} & {High} & {Low} & {Med.} & {High} \\ 
			\midrule
			Ground truth  & 1.00 & 1.00 & 1.00 & 1.00 & 1.00 & 1.00 \\
			Unadjusted & 1.03 & 1.24 & 3.48 & \maxf{0.99} & 1.22 & 3.51 \\ 
			\cmidrule{1-7} 
			NN \qhat & 1.03 & 1.18 & 2.04 & 0.89 & 1.08 & 2.24 \\
			NN \plugin & 1.03 & 1.18 & 1.40 & 0.85 & 1.05 & 2.07 \\
			 C-ATM \qhat  & \maxf{1.01} & 1.16 & 2.45 & 1.04 & \maxf{1.04 }& 1.72 \\
			 C-ATM \plugin  & 1.01 & 1.13 & 2.09 & 0.95 & 0.94 & \maxf{1.11 }\\
			 C-BERT \qhat & 1.07 & \maxf{1.07 }& 1.14 & 1.50 & 0.95 & 1.12\\
			 C-BERT \plugin & 1.08 & 1.15 &\maxf{0.94} & 2.07 & 1.07 & 1.27 \\
			\bottomrule
		\end{tabular}
	\end{center}
\end{table*}

\begin{table*}[h!t]
	\caption{Embedding adjustment recovers the ATT on PeerRead.
		This persists even with high confounding.
		Table entries are estimated ATT.
		Columns are labeled by confounding level.
		Low, Med., and High correspond to $\beta_1=1.0,5.0$ and $25.0$.  }\label{tb:simPeerRead}
	\begin{center}
		\begin{tabular}{l @{\hskip 3em} *{3}{S}}
			\toprule
			\multicolumn{1}{r}{Confounding:\ \ }  & {Low} & {Med.} & {High} \\ 
			\midrule
			Ground truth  & 0.06 & 0.05 & 0.03  \\
			Unadjusted & 0.08 & 0.15 & 0.16  \\
			\cmidrule{1-4}
			NN \qhat & \maxf{0.05} & 0.10 & 0.30 \\
			NN \plugin & 0.05 & 0.10 & 0.30 \\
			C-ATM \qhat & 0.07 & 0.10 & 0.32 \\
			C-ATM \plugin & 0.07 & 0.10 & 0.32 \\
			C-BERT \qhat & 0.09 & \maxf{0.07} & \maxf{0.04} \\
			C-BERT \plugin & 0.10 & 0.09 & 0.05 \\
			\bottomrule
		\end{tabular}
	\end{center}
\end{table*}

\begin{table*}[h!]
	\caption{Embedding adjustment reduces estimated treatment effects in PeerRead.
		Entries are estimated treatment effect and 10-fold bootstrap standard deviation.
	}\label{tb:peerread}
	\begin{center}
		\begin{tabular}{l @{\hskip 0em} *{2}{S} }
			\toprule
			{} & {\texttt{buzzy}} & \texttt{theorem} \\ 
			\midrule
			Unadjusted  & 0.08\pm0.01  & 0.21\pm0.01 \\
			C-BERT \qhat & -0.03\pm0.01 & 0.16\pm0.01  \\
			C-BERT \plugin & -0.02\pm0.01 & 0.18\pm0.02 \\
			\bottomrule
		\end{tabular}
	\end{center}
\end{table*}
\begin{table*}[h!]
	\caption{Embedding adjustment reduces estimated direct effects in Reddit.
		Entries are estimated treatment effect and 10-fold bootstrap standard deviation.
	}\label{tb:Reddit}
	\begin{center}
		\begin{tabular}{l @{\hskip 0em} *{3}{S} }
			\toprule
			{} & \texttt{okcupid} & \texttt{childfree} & \texttt{keto} \\ 
			\midrule
			Unadjusted & -0.18\pm0.01 & -0.19\pm0.01 & -0.00\pm0.00  \\
			C-BERT \qhat & -0.10\pm0.04 & -0.10\pm0.04  & -0.03\pm0.02  \\
			C-BERT \plugin & -0.15\pm0.05 & -0.16\pm0.05 &
			-0.01\pm0.00  \\
			\bottomrule
		\end{tabular}
	\end{center}
\end{table*}
\begin{figure}[th!]
	\captionsetup{width=0.5\textwidth}
	\label{fig:Reddit-exogeneity}
	\begin{center}
		\includegraphics[width=0.5\textwidth]{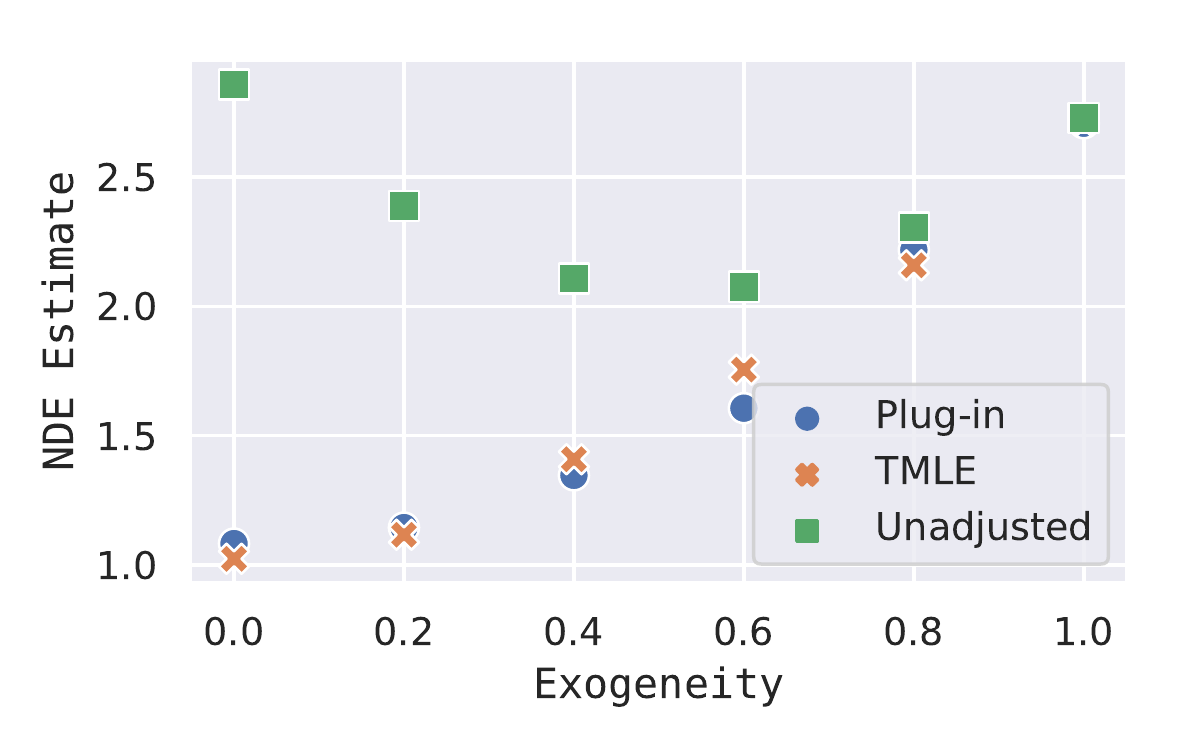}
	\end{center}
	\caption{The method improves the unadjusted estimator even with
		exogeneous mediatiors. Plot shows estimates of NDE from simulated
		data based on Reddit.  Ground truth is 1.}
\end{figure}

\subsection{Results}
Semi-synthetic data is used to investigate three questions about
causal adjustment:
1) does language modeling help? 2) does supervised dimensionality reduction help?
3) do C-BERT and C-ATM produce embeddings that sufficiently adjust for confounding?
Results are summarized in \cref{tb:q1andq2,tb:simReddit,tb:simPeerRead}.

\parhead{Language modeling helps.}
The left table in \cref{tb:q1andq2} illustrates the point that 
representations that model language produce better causal estimates.
C-ATM and C-BERT recover the simulated treatment effect in Reddit
and PeerRead more accurately than NN and BERT (sup. only) .
Among them, C-BERT performs best.
This experiment uses the confounding setting $\beta_1=10.0, \gamma=1.0$ for Reddit
and $\beta_1=5.0$ for PeerRead. 

\parhead{Supervision helps.}
The right table in \cref{tb:q1andq2} summarizes effect estimation
for fixed representations that are used to fit expected outcomes and propensity scores.
Among these, LDA, ATM and BERT model language structure.
No method estimates the treatment effects as accurately
as C-BERT or C-ATM.
The finding suggests supervising representations to discard information not relevant
to predicting the treatment or outcome is useful for effect estimation.
The improvement of C-ATM over ATM and C-BERT over BERT suggests that
combining language modeling with supervision works best.

\parhead{Methods adjust for confounding.}
\cref{tb:simReddit,tb:simPeerRead} show the quality of effect estimation
as the amount of confounding increases. 
For Reddit, the confounding setting $\beta_1$ varies across $1.0, 10.0, 100.0$ and
the noise $\gamma$ varies across 1.0 and 4.0.
For PeerRead, $\beta_1$ varies across $1.0, 5.0, 25.0$.
Results from the NN method are
shown as a baseline.

Compared to the unadjusted estimate,
all methods
perform adjustments that 
reduce confounding.
However, C-BERT and C-ATM recover the most accurate
causal estimate in all settings.  
In the Reddit setting, C-ATM performs best when the outcome variance is higher ($\gamma=4.0$).
However, C-BERT is best when the outcome variance is lower ($\gamma=1.0$).
In the PeerRead setting, C-BERT performs best.

\parhead{The effect of exogeneity.}
We assume that the text carries all information about the confounding (or mediation) necessary to identify the causal effect.
In many situations, this assumption may not be fully realistic.
For example, in the simulations just discussed, it may not be possible to exactly recover
the confounding from the text.
We study the effect of violating this assumption by simulating both treatment and outcome
from a confounder that consists of a part that can be fully inferred from the text and part that is wholly exogenous.

  The challenge is finding a realistic confounder that can be exactly inferred from the text.
Our approach is to (i) train BERT to predict the actual treatment of interest, producing
propensity scores $\hat{g}_i$ for each $i$, and (ii) use $\hat{g}_i$ as the inferrable part of the confounding.
Precisely, we simulate propensity scores as $\logit g_{\text{sim}} = (1-p)\logit\hat{g}_i + p \xi_i$,
with $\xi_i \distiid \mathrm{N}(0,1)$. The outcome is simulated as above.

When $p=0$, the simulation is fully-inferrable and closely matches real data.
Increasing $p$ allows us to study the effect of exogeneity; see \cref{fig:Reddit-exogeneity}.
As expected, the adjustment quality decays.
Remarkably, the adjustment improves the naive estimate at all levels of exogeneity---the
method is robust to violations of the theoretical assumptions.

\parhead{Application}
We apply causal BERT to estimate the treatment effect of \texttt{buzzy} and \texttt{theorem},
and the effect of gender on log-score in each subreddit. 
Although unadjusted estimates suggest strong effects,
our results show this is in large part explainable by confounding or mediating.
See \cref{tb:peerread,tb:Reddit}.
On PeerRead, both estimates
suggest a positive effect from including a theorem
on paper acceptance.
On Reddit, both estimates suggest a positive effect from labeling a post as female 
on its score in \texttt{okcupid} and \texttt{childfree}.

\section{DISCUSSION}
We have examined the use of black box embedding methods for causal inference with text.
The challenge is to produce a low-dimensional representation of text 
that is sufficient for causal adjustment.
We adapt two modern tools for language modeling --- BERT and topic modeling ---
to produce representations that predict the treatment and outcome well.
This marries two ideas to produce causally sufficient text representations:
modeling language and supervision relevant to causal adjustment.
We propose a methodology for empirical evaluation that uses real text documents
to simulate outcomes with confounding.
The studies in this simulated setting validate the representation learning
insights of this paper. The application to real data completes the demonstration
of our methods.

There are several directions for future work.
First, the black box nature of the embedding methods makes it difficult for practitioners to
assess whether the causal assumptions hold: is it possible to develop visualizations and sensitivity analyses
to aid these judgments?
Second, we require both the treatment and outcome to be external to the text.
How can the approach here be extended to estimate the causal effect of (or on) aspects of the writing itself?
Third, the deep learning methods we have used are mainly geared towards predictive performance.
Are there improvements that will help with estimation accuracy?
For example, should we adapt methods that specifically target well-calibrated predictions?
Finally, how can methods in this domain be reliably empirically evaluated?
We have developed a new strategy for realistic semi-synthetic simulations in this paper.
Can our approach be extended to a complete approach for benchmarking?

\section{ACKNOWLEDGEMENTS}
This work was supported by ONR N00014-15-1-2209,
ONR N00014-17-1-2131 , NIH 1U01MH115727-01, NSF CCF-1740833, DARPA SD2 FA8750-18-C-0130,
the Alfred P. Sloan Foundation, 2Sigma, and the government of Canada through NSERC. The GPUs used
for this research were donated by the NVIDIA Corporation.

\printbibliography

\end{document}

